\ifdefined\pdfminorversion\pdfminorversion=7\fi
\documentclass[letterpaper, 10 pt, conference]{ieeeconf}

\IEEEoverridecommandlockouts
\usepackage{amsmath,amssymb}
\usepackage{booktabs}
\usepackage{graphicx}
\usepackage{adjustbox}
\usepackage{multirow}
\usepackage{xcolor}
\usepackage{url}
\usepackage{cite}
\usepackage{microtype}
\usepackage{placeins}
\usepackage{stfloats}

\newcommand{\tabletitle}[1]{%
  \refstepcounter{table}%
  \par\begingroup\small\leftskip=0pt\rightskip=0pt
  \parfillskip=0pt plus 1fil\noindent
  \textbf{TABLE~\Roman{table}:}~#1\par\endgroup\vspace{2pt}}

\AtBeginDocument{%
  \setlength{\abovedisplayskip}{2pt plus 1pt minus 1pt}%
  \setlength{\belowdisplayskip}{2pt plus 1pt minus 1pt}%
  \setlength{\abovedisplayshortskip}{1pt plus 1pt}%
  \setlength{\belowdisplayshortskip}{2pt plus 1pt minus 1pt}%
}

\makeatletter
\renewcommand\section{\@startsection{section}{1}{\z@}
  {6pt}{3pt}{\normalfont\normalsize\centering\scshape}}
\renewcommand\subsection{\@startsection{subsection}{2}{\z@}
  {6pt}{3pt}{\normalfont\normalsize\itshape}}
\makeatother

\title{\LARGE \bf
Touch Where It Matters: Learning Local Tactile\\
Policies for Fine-Grained Gripper Control}

\author{%
\authorblockN{Mingyuan Ba, Yanan Zhou, James Zhao, and Weiming Zhi$^{*}$}
\authorblockA{School of Computer Science,\\
The University of Sydney, Australia\\
$^{*}$Corresponding author: \texttt{weiming.zhi@sydney.edu.au}}
}

\begin{document}
\bstctlcite{IEEEbstControl}

\flushbottom
\maketitle
\thispagestyle{empty}
\pagestyle{empty}

\begin{abstract}
Visual imitation policies can organise task motion, yet contact-rich object
use demands gripper adjustments beyond coarse opening and closing. We study
fine-grained gripper control as a local feedback problem within ongoing
task execution. Our approach assigns tactile feedback a dedicated control
role during contact-rich task stages. We introduce ContactPolicy, a learned
local controller that takes over the gripper while a visual base policy
continues to control the arm. ContactPolicy combines tactile and force
histories with current robot state to jointly predict gripper increments and
their modulation, updating the gripper target from each new observation. An
Intent policy manages control handover between the visual base policy and ContactPolicy.
Real-world experiments on container opening, object handling, and tool
operation show improved task success over both the Visuomotor Policy and
Tactile--Force Policy baselines. Execution analysis illustrates how the gripper
adjusts contact within an ongoing task, including periods of target
retention and fine-grained changes as interaction conditions evolve.
Arm-state ablations and failure analysis further examine how execution
context supports local regulation. These results support task-local tactile
policies for fine-grained gripper control within visual manipulation.
\end{abstract}

\suppressfloats[t]
\begin{figure}[t]
    \centering
    \includegraphics[width=\linewidth]{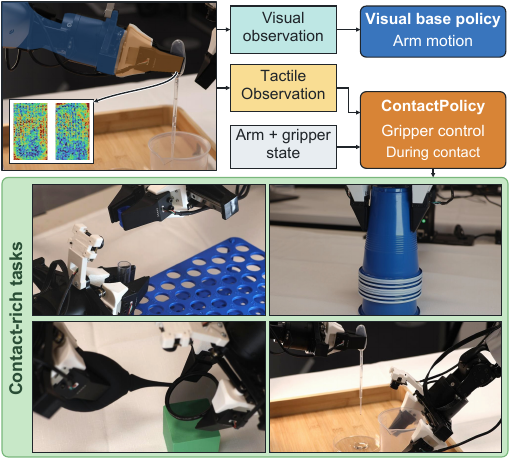}
    \caption{\textbf{Touch where it matters.} Visual arm control with local tactile gripper regulation. Photos and tactile examples are not time-synchronised.}
    \label{fig:teaser}
\end{figure}

\section{Introduction}

Visual imitation learning has enabled robots to acquire increasingly
complex manipulation skills from demonstrations. Visual imitation policies
learn coordinated action sequences from visual observations and robot
state~\cite{chi2023diffusion}, while Mobile ALOHA and UMI extend
data collection to bimanual, mobile, and in-the-wild
settings~\cite{fu2025mobile,chi2024umi}. Localisation feedback can improve
mobile execution by tracking map-frame base-pose targets, as in
MAVP~\cite{tang2026mavp}. Alongside reliable task motion, success also
depends on how the gripper interacts with the object during contact.

In the visual-policy executions we study, gripper behaviour is often
dominated by coarse opening and closing. Such commands can acquire and
release objects, but provide limited control over their use. A loaded
dropper must be held during transport and squeezed to dispense; a blower
must be compressed while the arm positions its outlet; and nested cups
must remain securely grasped as the arm separates them. Here, the gripper
actively performs the operation. Even when arm motion is appropriate,
small errors in gripper adjustment can determine whether the task succeeds.
Visual runtime monitors such as PATCH detect persistent local scene
disturbances and trigger intervention~\cite{zhou2026patch}. Object use
also requires feedback that regulates the ongoing gripper--object interaction.

These tasks expose a local control problem within visual manipulation:
\emph{when should the gripper change its target, by how much, and when
should it hold that target?} Touch is well suited to these decisions
because it senses the contact interface where gripper actions take
effect. Visual--tactile datasets and tactile skill learning demonstrate
the value of this complementary
information~\cite{yang2022touchandgo,fu2024tvl,guzey2024seetotouch,lee2024dextouch,lin2025hato}.
Contact history reveals how an interaction evolves, while robot state
provides the context needed to coordinate adjustments with arm motion.

We therefore study \emph{task-local fine-grained gripper control}
(Fig.~\ref{fig:teaser}): tactile feedback regulates the gripper during
contact-rich stages while the visual base policy continues to advance the arm.
The relevant feedback is concentrated at the fingertips and is most useful
during object use. Assigning it to this stage lets the gripper respond to
changing contact without requiring the visual base policy to resolve every small
adjustment.
Recent reactive, force-aware, and modular policies explore complementary
ways to incorporate contact feedback~\cite{xue2025rdp,he2025foar,helmut2025farm,zhang2026touchguide,chen2026policyconsensus},
and RETAF separates arm prediction from gripper force
regulation~\cite{kang2026retaf}. Our focus is learning gripper position
updates together with the decision to apply or withhold each update.
Locality specifies both the controlled component and the stage in which
contact feedback assumes control.

We introduce ContactPolicy to implement this division of control. It
encodes tactile and force histories, combines them with current arm and
gripper states, and jointly predicts gripper increments and their
modulation. The controller applies the first modulated increment to the
commanded target, then replans from the next observation. An Intent
policy manages handover between the visual base policy and ContactPolicy,
allowing local contact regulation within ongoing task execution.

Real-robot experiments improve task success over the Visuomotor Policy
and Tactile--Force Policy baselines. Execution analyses highlight how local
contact adjustment coordinates gripper operation with arm motion.

Our contributions are as follows:
\begin{enumerate}[\topsep=2pt\partopsep=0pt\parsep=0pt\itemsep=0pt]
    \item We formulate task-local fine-grained gripper control for
    contact-rich manipulation: the arm advances the task while the gripper
    uses local feedback to adjust contact beyond coarse opening and closing.
    \item We introduce ContactPolicy, a learned local tactile controller
    that combines contact histories and current robot state to predict and
    modulate gripper increments. Control handover integrates its iterative
    target updates into visual task execution.
    \item We evaluate ContactPolicy on real-world contact-rich
    tasks, demonstrating improved success over the visual base policy.
    Input ablations and execution failure analysis examine how robot state
    supports local tactile regulation and coordination with task motion.
\end{enumerate}

\section{Related Work}

\noindent\textbf{Visual imitation and generalist policies.}
ACT and Diffusion Policy learn action sequences from
 demonstrations~\cite{zhao2023act,chi2023diffusion}; Mobile ALOHA and UMI
extend demonstration collection to mobile and in-the-wild
settings~\cite{fu2025mobile,chi2024umi}. Learned hand skills can assist
dexterous demonstration collection~\cite{zhao2026nestdex}.
Open X-Embodiment and DROID expand
robot experience~\cite{oxe2024,khazatsky2024droid}, while Octo and OpenVLA
learn generalist policies across tasks and
environments~\cite{ghosh2024octo,kim2025openvla}.
Demonstration timing also affects learned coordination: retimed segments
support synchronous multi-arm policy learning~\cite{zhao2026trimanpolicy}.
Our focus is the contact adjustment required within visual task execution.

\noindent\textbf{Tactile representations and multimodal learning.}
UniTouch aligns touch with other modalities~\cite{yang2024unitouch}.
Sparsh, T3, and UniT learn reusable contact features across tasks,
sensors, or objects~\cite{higuera2024sparsh,zhao2024t3,xu2025unit}.
AnyTouch and AnyTouch 2 model static and dynamic tactile information
across sensors~\cite{feng2025anytouch,feng2026anytouch2}, while Tactile
Beyond Pixels combines complementary touch
channels~\cite{higuera2025tactilebeyondpixels}. These advances support
contact perception and downstream manipulation. A reusable representation
must still be mapped to task-appropriate actions during closed-loop
execution. We study the control
role of contact feedback, using its history and current robot state
to predict and modulate local gripper increments from updated contact observations.

\noindent\textbf{Reactive tactile and force-aware control.}
RDP and FoAR couple visual task execution with reactive contact
feedback~\cite{xue2025rdp,he2025foar}. TacDiffusion and FARM learn
force-aware actions~\cite{wu2025tacdiffusion,helmut2025farm}, while RETAF
decouples arm prediction and gripper force regulation~\cite{kang2026retaf}.
TouchGuide guides action sampling with a tactile model, while Policy
Consensus combines modality-specific experts through learned
routing~\cite{zhang2026touchguide,chen2026policyconsensus}.
At the team level, CoRE routes collaboration-role experts within a shared
policy for decentralized multi-robot manipulation~\cite{zhou2026core}.
ContactPolicy focuses on task-local gripper position updates: contact
history and execution state condition increments and demonstrated-action
modulation, with Intent managing control handover. Ablations assess
execution context for local regulation within the ongoing visual task execution.
Each update responds to the evolving contact.

\section{ContactPolicy: Local Tactile Gripper Control}

\subsection{Problem Formulation}

We study manipulation in which an object must be actively operated after
it is grasped. A visual base policy $\pi_{\mathrm{high}}$ maps observation
$o_t$ to arm actions and a nominal gripper command. During a contact-rich
stage, the arm advances the task while the gripper makes fine-grained
width adjustments to operate the grasped object. We define
\emph{task-local fine-grained gripper control} as learning these adjustments
from contact feedback within the ongoing visual execution.

Let $g_t\in[g_{\min},g_{\max}]$ be the commanded gripper width before an
update, and let $o_t^c$ comprise recent tactile and force observations and
the current arm and gripper states. Given synchronised demonstrations of
these observations and commanded widths, the objective is to learn a local
policy that decides whether to update $g_t$ and predicts an increment
$\Delta g_t$, with $\Delta g_t=0$ retaining the target. Thus, target retention is part of the learned control behaviour.

Control is local in both component and stage: the learned policy assumes
gripper control during active contact operation. The visual base policy continues arm
control, while measured arm state conditions local adjustments. Stage
selection determines when the local policy takes control; target updates
determine how it acts within that stage.

\subsection{System Overview}

\begin{figure}[!h]
    \centering
    \includegraphics[width=\linewidth]{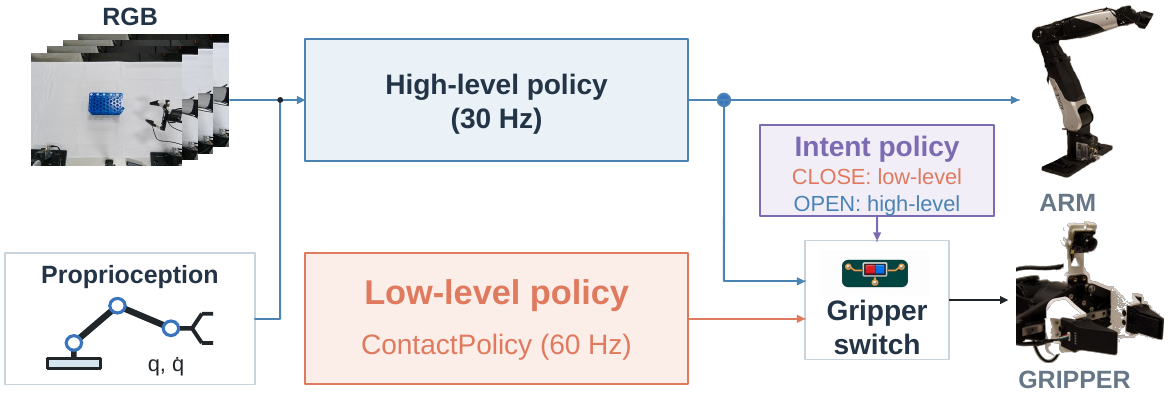}
    \caption{\textbf{System overview.} The visual base policy advances the arm; Intent assigns gripper control to the visual base policy or ContactPolicy.}
    \label{fig:method_system}
\end{figure}

\begin{figure*}[!t]
    \centering
    \includegraphics[width=\linewidth]{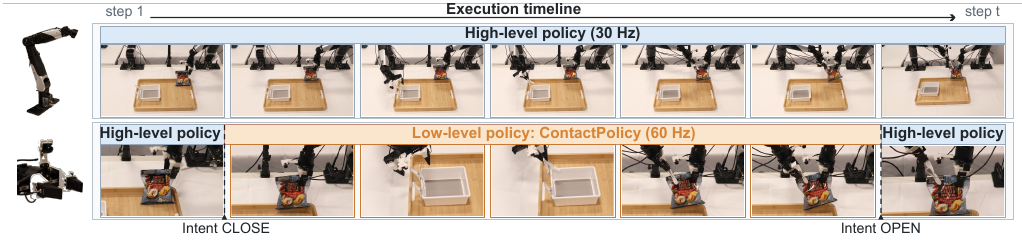}
    \caption{\textbf{Gripper control handover.} The visual base policy continues arm motion while Intent CLOSE activates local gripper regulation. Intent OPEN returns gripper control to the visual base policy after the operation, linking the local feedback stage to the surrounding task execution.}
    \label{fig:method_execution}
\end{figure*}

As shown in Fig.~\ref{fig:method_system}, ContactPolicy implements this
control assignment through three components. The visual base policy provides arm motion and nominal gripper commands. ContactPolicy uses
contact history and measured robot state to update the gripper target.
Intent selects which policy supplies the gripper command. Let
$\mu_t\in\{\mathrm{high},\mathrm{low}\}$ denote this selection:
\begin{equation}
 u_t^{g} =
 \begin{cases}
 \pi_{\mathrm{low}}(o_t^c,g_t), & \mu_t=\mathrm{low},\\
 \pi_{\mathrm{high}}^{g}(o_t), & \mu_t=\mathrm{high}.
 \end{cases}
 \label{eq:hierarchical_control}
\end{equation}
The local policy predicts an increment and a modulation decision, then
converts their product into a gripper target. Intent determines
\emph{who controls the gripper}; local modulation determines
\emph{whether to apply the predicted adjustment}.

In the bimanual system, each arm runs an independently parameterised
ContactPolicy conditioned on its own tactile, force, arm-state, and
gripper-state observations. Both models share a deployment process and
Intent inference, with local feedback computed independently for each arm.

\begin{samepage}
\noindent\textbf{Control handover.}
Fig.~\ref{fig:method_execution} illustrates entry into and exit from local
control while arm execution continues. Intent uses a sequence model to predict
an OPEN/CLOSE score $c_t^{\mathrm{intent}}=\mathcal{I}(o_t^{\mathrm{intent}})$
for each gripper.
\par
\end{samepage}
The gripper output slots are trained with OPEN=0 and CLOSE=1 labels derived
from demonstration width trajectories, as described in the training
section. These labels supervise control ownership at the task-stage level,
complementing the update decisions learned within ContactPolicy.

A consecutive-prediction counter confirms a switch when scores remain
on the same side of the Intent threshold over a confirmation window. Confirmed CLOSE activates ContactPolicy; confirmed OPEN restores the
visual base policy's gripper output. Throughout this process, measured arm state
provides execution context for local adjustments while the visual base policy
maintains control of the arm.

\noindent\textbf{Intent observations.}
Intent combines RGB views with joint-and-gripper state, using both arm
views and the top view for bimanual execution, and the right and top views
for right-arm execution. An ImageNet-pretrained
ResNet backbone and a transformer map these observations to the
control-ownership signal. This visual context places each handover within
the surrounding task while contact feedback drives local target updates.

\subsection{Learning Local Tactile Feedback}

\begin{figure}[!t]
    \centering
    \includegraphics[width=\linewidth]{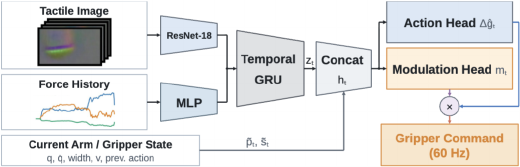}
    \caption{\textbf{Local tactile policy.} Contact histories and current robot state condition an increment $\Delta\hat g_t$ and modulation decision $m_t$. Their product enters the gripper-command stage, which updates the target at 60\,Hz.}
    \label{fig:method_contact}
\end{figure}

\noindent\textbf{Combining Contact History with Execution State.}
A gripper adjustment takes effect within ongoing object motion.
ContactPolicy (Fig.~\ref{fig:method_contact}) combines contact history,
which describes how the
interaction evolves, with current robot state, which supplies its execution
context. Together, they condition local adjustments on both fingertip
feedback and task motion.

The local observation is
$o_t^c=(I^0_{t-H+1:t},I^1_{t-H+1:t},F_{t-H+1:t},p_t,s_t)$.
The two tactile streams $I^0,I^1$ provide fingertip RGB difference images,
and $F_t\in\mathbb{R}^6$ concatenates the three force components from each
fingertip. Arm state $p_t=[q_t,\dot q_t]\in\mathbb{R}^{14}$ contains the
seven joint positions and velocities. Velocities are taken from the robot
interface or estimated from consecutive position measurements. Gripper
state $s_t$ contains measured width, width velocity, and the previous
applied command increment, $\Delta g_{t-1}^{\mathrm{app}}=g_t-g_{t-1}$.
This last input relates the observed contact to the controller's preceding
change in target. Demonstration inputs use the corresponding difference
between consecutive commanded widths.

At each history step $\tau$, a ResNet-18 image encoder extracts tactile
image features and a multilayer perceptron encodes force. Their concatenation
$x_\tau$ is aggregated over $H$ observations by a gated recurrent unit (GRU)~\cite{cho2014gru}:
\begin{equation}
\begin{aligned}
z_t&=\operatorname{GRU}_{\mathrm{contact}}(x_{t-H+1:t}),\\
h_t&=[z_t,\widetilde p_t,\widetilde s_t].
\end{aligned}
\label{eq:contact_encoding}
\end{equation}

Current normalised arm and gripper states join the contact representation
after temporal aggregation, providing both prediction heads with the
latest execution context. Each
inference call starts a fresh GRU computation on the explicit window;
short histories are padded by repeating the earliest valid observation.

\noindent\textbf{Observation preparation.}
We resize Xense driver-provided RGB difference images to the encoder
input resolution and scale their intensities to the unit interval. The six force channels
represent fingertip forces in newtons under the driver convention.
Force inputs are standardised and robot states normalised using
training-set statistics. The wrist-camera variant adds features from a separate
encoder for the corresponding arm's RGB images, using the same image
preprocessing. It extends the local temporal representation while
preserving Intent's observation interface and switching rule. The
arm-state ablation in Sec.~\ref{sec:arm_state_ablation} evaluates how current
motion context contributes to local regulation.

\begin{figure}[!t]
    \centering
    \includegraphics[width=\linewidth]{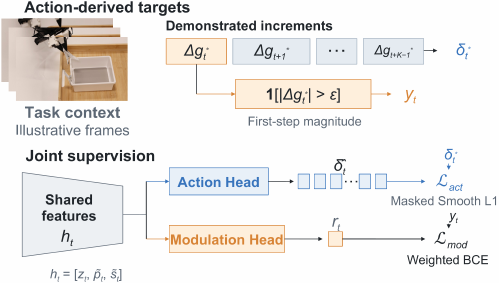}
    \caption{\textbf{Action and modulation supervision.} Demonstrated increments train predictions, while update/hold labels supervise the modulation head.}
    \label{fig:action_supervision}
\end{figure}

\noindent\textbf{Learning When and How Much to Adjust.}
Fine-grained object use calls for both changing the gripper configuration
and holding a useful target as the task progresses. ContactPolicy learns
these complementary decisions through an action head and a modulation
head. The action head predicts a $K$-step sequence of physical width
increments. The modulation head produces an update probability $r_t$ and
its thresholded execution output $m_t$:
\begin{equation}
\begin{aligned}
\widehat{\boldsymbol{\delta}}_t&=\pi_\theta(h_t),
\quad r_t=\sigma(f_{\mathrm{mod}}(h_t)),\\
m_t&=\mathbb{1}[r_t\geq\tau_{\mathrm{mod}}].
\end{aligned}
\label{eq:action_prediction}
\end{equation}
Here, $\sigma$ is the sigmoid function and $\tau_{\mathrm{mod}}$ is the
update threshold. Fig.~\ref{fig:method_contact} shows the binary output
$m_t$ used during execution; $r_t$ supplies the training loss in
Fig.~\ref{fig:action_supervision}. A tanh-based physical action
mapping converts the action head's unconstrained outputs into increments. Positive values open the gripper;
negative values close it. The predicted sequence provides a local action
horizon, and each new observation triggers replanning. Execution uses its
first increment, $\Delta\hat g_t=\widehat{\boldsymbol{\delta}}_t[1]$,
expressed in metres per training frame.

\noindent\textbf{Feedback-driven target updates.}
Following Fig.~\ref{fig:method_contact}, the two head outputs are multiplied
to form a modulated increment. The gripper-command stage clips this
increment to bounds $\delta_{\min}\leq0\leq\delta_{\max}$ and compensates for the training and
execution rates:
\begin{equation}
\begin{aligned}
\Delta g_t^{\mathrm{mod}}&=m_t\Delta\hat g_t,\\
\Delta g_t^{\mathrm{cand}}&=
\frac{f_{\mathrm{train}}}{f_{\mathrm{exec}}}
\operatorname{clip}(\Delta g_t^{\mathrm{mod}},\delta_{\min},\delta_{\max}).
\end{aligned}
\label{eq:execution_candidate}
\end{equation}
A physical deadband removes candidates at or below the minimum adjustment
magnitude $\epsilon$. The remaining increment updates the commanded width
within the gripper's target bounds:
\begin{equation}
\begin{aligned}
\Delta g_t&=\mathbb{1}[|\Delta g_t^{\mathrm{cand}}|>\epsilon]
\Delta g_t^{\mathrm{cand}},\\
g_{t+1}&=\operatorname{clip}(g_t+\Delta g_t,g_{\min},g_{\max}).
\end{aligned}
\label{eq:target_accumulation}
\end{equation}
Together, these operations define the 60\,Hz Gripper Command block in
Fig.~\ref{fig:method_contact}, with $u_t^g=g_{t+1}$ during local control.
When $m_t=0$, the modulated increment is zero and the current target is
retained while sensing and prediction continue. After target clipping, the applied
increment $\Delta g_t^{\mathrm{app}}=g_{t+1}-g_t$ is recorded for the next
observation. Feeding back this executed command difference keeps the state
input consistent with target saturation. The controller therefore learns
contact-dependent adjustments and implements them through successive
updates to a persistent gripper target.

\begin{figure*}[!t]
    \centering
    \includegraphics[width=\linewidth]{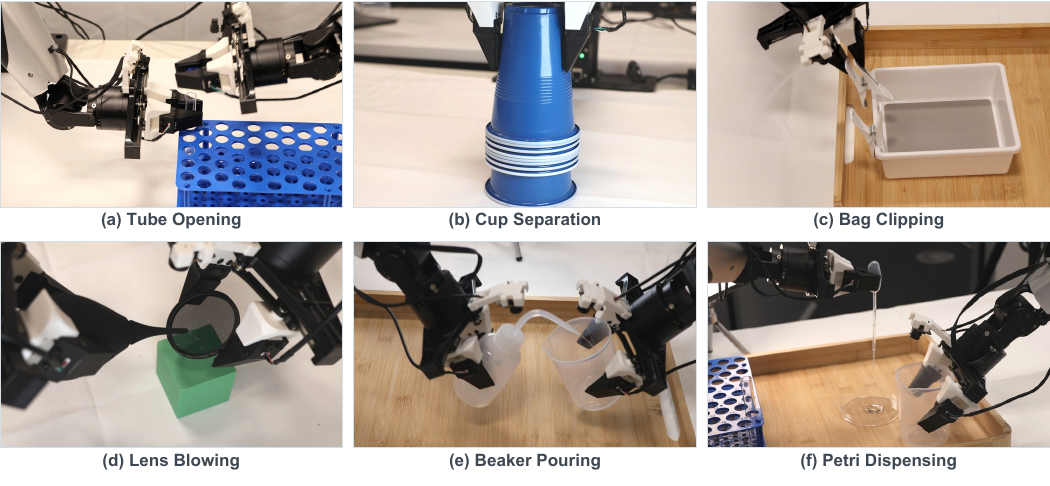}
    \caption{\textbf{Six contact-rich manipulation tasks.} (a) Tube Opening, (b) Cup Separation, (c) Bag Clipping, (d) Lens Blowing, (e) Beaker Pouring, and (f) Petri Dispensing test fine-grained gripper control during container opening, object handling, and tool operation, with contact feedback during task motion.}
    \label{fig:task_overview}
\end{figure*}

\subsection{Training and Deployment}

\noindent\textbf{Supervising Adjustment and Retention.}
Synchronised demonstrations provide tactile and force observations, robot
states, and commanded gripper widths. They supervise both the magnitude
of an adjustment and the decision to make one, as illustrated in
Fig.~\ref{fig:action_supervision}. For each $K$-step action horizon, adjacent
commanded widths define the increment targets, and the first increment
defines the update label:
\begin{equation}
\begin{aligned}
\delta_{t,k}^\star&=g_{t+k}^{\star,\mathrm{cmd}}
-g_{t+k-1}^{\star,\mathrm{cmd}},\quad k=1,\ldots,K,\\
y_t&=\mathbb{1}[|\delta_{t,1}^\star|>\epsilon].
\end{aligned}
\label{eq:demonstration_increment}
\end{equation}
We denote the target sequence by $\boldsymbol\delta_t^\star$ and write
$\Delta g_t^\star=\delta_{t,1}^\star$. Update labels use physical increments
before normalisation. Let $\mathcal N$ denote training-set min--max action
normalisation. The two heads are trained with
\begin{equation}
\begin{aligned}
\mathcal L_{\mathrm{act}}&=
\mathbb E_{t:y_t=1}\!\left[
\operatorname{SmoothL1}\!\left(
\mathcal N(\widehat{\boldsymbol\delta}_t),
\mathcal N(\boldsymbol\delta_t^\star)\right)\right],\\
\mathcal L_{\mathrm{mod}}&=\operatorname{BCE}_{w_+}(r_t,y_t).
\end{aligned}
\label{eq:action_loss}
\end{equation}
Action regression trains the full predicted sequence on windows whose
first demonstrated step is an update. Modulation training uses both update
and hold samples, with positive-class weight $w_+$ set to the batch's
negative-to-positive ratio and clipped to a bounded range. Together, the two losses teach the magnitude and timing of local target
updates from the same demonstrations.

Auxiliary branches supervise the contact representation by predicting
future tactile features and the force vector over a short prediction
horizon. Tactile feature targets come from an exponential-moving-average
encoder. Force prediction is
supervised in newtons, complementing the standardised force input.
The complete objective combines these terms:
\begin{equation}
\begin{aligned}
\mathcal L={}&\mathcal L_{\mathrm{act}}
+\lambda_{\mathrm{mod}}\mathcal L_{\mathrm{mod}}\\
&+\lambda_{\mathrm{latent}}\mathcal L_{\mathrm{latent}}
+\lambda_{\mathrm{force}}\mathcal L_{\mathrm{force}},
\end{aligned}
\label{eq:total_loss}
\end{equation}
The auxiliary branches shape the representation during training; deployment
uses the contact encoder and the action and modulation heads.

\noindent\textbf{Integrating the Learned Controller.}
Intent is trained separately from ContactPolicy using labels derived from
demonstrated width trajectories. Median filtering precedes a direction
test with a physical deadband. Sustained changes in one direction switch
the label, followed by a minimum retention period. Within the deadband, the preceding label persists, yielding a handover
signal from demonstrated opening and closing behaviour.

A confirmed handover resets observation history, the previous applied
increment, and the target accumulator. The next active inference anchors
the target to the measured gripper width before resuming local feedback. Each
subsequent step follows
Eqs.~\eqref{eq:execution_candidate}--\eqref{eq:target_accumulation}, using the
previous clipped target as its starting command. Local commands follow the
modulated-increment update directly, independently of the high-level
smoother and step limiter. The final applied target difference becomes
part of the next state observation.

The visual base policy continues arm control and retains its action chunk during local
regulation. Confirmed OPEN restores its gripper output through a guard
using the previous command, exponential moving averaging, and a
maximum-step limit. The resulting interface separates control ownership
from local update timing, allowing the tactile policy to retain or adjust
the gripper target as the arm advances.

\section{Experiments}
\label{sec:experiments}

We evaluate local gripper control through task success, contact
behaviour, arm coordination, observation design, and object outcomes
to connect contact-stage control with successful object use during arm motion.

\begin{table}[t]
    \centering
    \small
    \tabletitle{\textbf{Task success across six contact-rich tasks.} Visuomotor Policy has no tactile input; Tactile--Force Policy follows tactile input-token fusion~\cite{heng2026vitacformer}. Success rates (\%) and their unweighted mean (Avg.) are shown; bold marks the best per column. TO: Tube Opening; CS: Cup Separation; BC: Bag Clipping; LB: Lens Blowing; BP: Beaker Pouring; PD: Petri Dispensing.}
    \label{tab:main_results}
    \setlength{\tabcolsep}{2.5pt}
    \begin{adjustbox}{max width=\linewidth}
    \begin{tabular}{@{}lccccccc@{}}
        \toprule
        \textbf{Method}
        & \textbf{TO}
        & \textbf{CS}
        & \textbf{BC}
        & \textbf{LB}
        & \textbf{BP}
        & \textbf{PD}
        & \textbf{Avg.} \\
        \midrule
        Visuomotor Policy    & 52\% & 0\% & 72\% & 48\% & 36\% & 48\% & 42.7\% \\
        Tactile--Force Policy & 72\% & 20\% & 68\% & 72\% & 56\% & 56\% & 57.3\% \\
        \textbf{ContactPolicy} & \textbf{100\%} & \textbf{52\%} & \textbf{92\%} & \textbf{88\%} & \textbf{80\%} & \textbf{84\%} & \textbf{82.7\%} \\
        \bottomrule
    \end{tabular}
    \end{adjustbox}
\end{table}

\noindent\textbf{Platform.}
Our platform combines AgileX Robotics NERO seven-DoF arms with Xense
Robotics optical tactile sensors. Each arm has a 3\,kg payload, 580\,mm
reach, and $\pm0.1$\,mm repeatability. Each sensor has a
$17.5\times29.5$\,mm sensing area, 0.03\,mm spatial resolution, and
UVC connectivity. Its specified 30--150\,Hz sampling range describes
sensor capability; controller scheduling is given below.

\noindent\textbf{Tasks and comparisons.}
The six tasks in Fig.~\ref{fig:task_overview} cover opening, separation,
clip handling, and tool use. Five tasks use bimanual execution; the remaining
task uses the right arm only. We compare \emph{Visuomotor Policy}, the visual base policy without tactile
input; \emph{Tactile--Force Policy}, which augments the Visuomotor Policy
with contact observations; and \emph{ContactPolicy}, combining visual arm
execution with local tactile gripper control. Each task/configuration uses
25 trials.
Equal trial counts make task-average and pooled success identical.
Ablations reuse full-model trials; execution images illustrate behaviour
within these cohorts.

\noindent\textbf{Tactile--Force Policy.}
Our Tactile--Force Policy follows prior tactile input-token
fusion~\cite{heng2026vitacformer}, using tactile difference-image and
fingertip-force histories. The two image streams are encoded and temporally
aggregated into one tactile token. An MLP and GRU encode the
six-dimensional XYZ force history from both fingertips into one force
token. The state encoder jointly processes the tactile and force tokens
with the current RGB observations.

\noindent\textbf{Data and execution.}
We collect 50 demonstrations per task through teleoperation and use the
same teleoperated episodes across methods. ContactPolicy training windows
are restricted to the contact-stage segments between gripper closing and
subsequent reopening; its visual base policy reuses the ACT
checkpoint. Data partitions follow complete episodes. For ContactPolicy, observations and commanded widths
are aligned on a 60\,Hz grid with a 50\,ms nearest-timestamp tolerance;
unmatched samples are discarded. ContactPolicy conditions each gripper update on a
fixed-length sliding window of recent tactile and force observations.
Visuomotor Policy, Tactile--Force Policy, and Intent run at
30\,Hz, while the local ContactPolicy runs at 60\,Hz. The tactile
difference images and fingertip-force histories for
Tactile--Force Policy are sampled directly at 30\,Hz. Only ContactPolicy
uses the 60\,Hz alignment for its local feedback updates.

\begin{figure*}[!t]
    \centering
    \includegraphics[width=\linewidth,trim=0bp 1.5bp 0bp 0bp,clip]{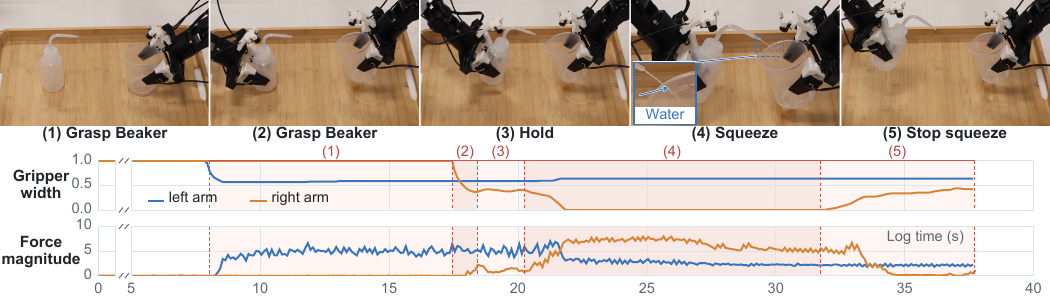}
    \caption{\textbf{Beaker Pouring phases and contact signals.} Gripper width (0--1) and force magnitude (N, 0.1\,s means) accompany representative task frames. Red markers indicate phases; the first 5\,s are compressed. Frames provide approximate phase correspondence; the inset enlarges the visible water stream.}
    \label{fig:beaker_contact_phases}
\end{figure*}

\begin{figure*}[!t]
    \centering
    \includegraphics[width=\linewidth]{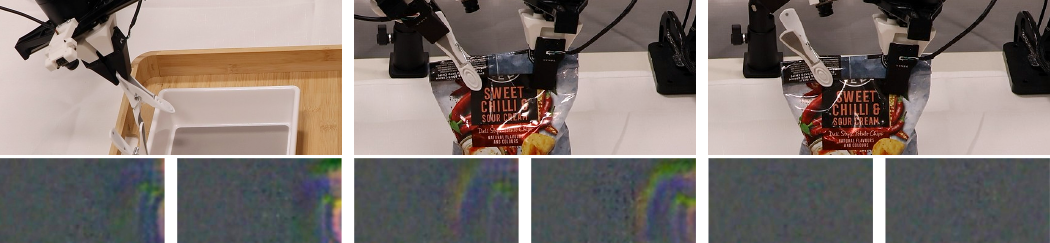}
    \caption{\textbf{Bag Clipping contact stages.} Frames show acquiring the clip, applying it to the bag, and releasing it. Paired tactile difference images from the gripper on the left reveal changing fingertip contact across these stages.}
    \label{fig:clip_contact_observations}
\end{figure*}

\subsection{Q1: Does Local Control Improve Task Success?}
\label{sec:task_performance}

ContactPolicy improves success on all six tasks in
Table~\ref{tab:main_results}, completing 124 of 150 trials for a mean success
rate of 82.7\%. It exceeds Visuomotor Policy and Tactile--Force Policy by
40.0 and 25.3 percentage points, respectively, with five tasks reaching at
least 80\% success. Evaluation covers complete object operations, including adjustments
made after grasp acquisition as the arm moves towards the task goal.

The two comparisons show the benefits of incorporating contact feedback
into task control. Fusing tactile and force histories into the visuomotor
policy raises mean success from 42.7\% to 57.3\%; ContactPolicy further
raises it to 82.7\% through local gripper regulation. ContactPolicy reuses
the ACT visual base policy checkpoint, leaving arm control with the visual
policy and assigning gripper control to the local policy during contact.
Its improvement over the contact-input fusion baseline on every task
supports assigning contact feedback to a local controller that adjusts the
gripper while the visual base policy continues to advance the arm.

These gains cover different object-operation requirements. Tube Opening,
Bag Clipping, and held-tool operation require further gripper adjustments
within established contact. Cup Separation requires selective release,
leaving exactly one cup on the table while lifting the remaining stack.
ContactPolicy achieves this outcome in 13/25 trials, compared with 0/25
for Visuomotor Policy and 5/25 for Tactile--Force Policy. The same division
of control therefore supports tool operation and relative motion between
objects, translating fine-grained contact adjustment into complete-task
success.

\subsection{Q2: How Does Local Feedback Shape Contact?}
\label{sec:contact_behaviour}

\begin{figure}[t]
    \centering
    \includegraphics[width=\linewidth]{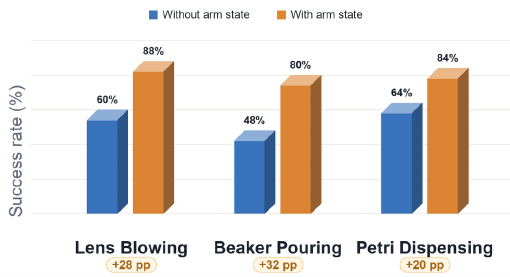}
    \caption{\textbf{Arm-state ablation.} Success over 25 trials per task improves with arm-state inputs on Lens Blowing, Beaker Pouring, and Petri Dispensing.}
    \label{fig:arm_state_ablation}
\par\vspace{3pt}
\begingroup
\centering
    \footnotesize
    \setlength{\tabcolsep}{1.5pt}
    \renewcommand{\arraystretch}{1.12}
    \tabletitle{\textbf{Arm-state ablation failures.} Total failures and premature actuation counts over 25 trials per task for each configuration.}
    \label{tab:failure_patterns}
    \begin{adjustbox}{max width=\linewidth}
    \begin{tabular}{@{}lcccccc@{}}
        \toprule
        \multirow{2}{*}{\textbf{Outcome}}
        & \multicolumn{2}{c}{\textbf{Lens Blowing}}
        & \multicolumn{2}{c}{\textbf{Beaker Pouring}}
        & \multicolumn{2}{c@{}}{\textbf{Petri Dispensing}} \\
        \cmidrule(lr){2-3}\cmidrule(lr){4-5}\cmidrule(l){6-7}
        & w/o & w/ & w/o & w/ & w/o & w/ \\
        \midrule
        Total failures & 10 & 3 & 13 & 5 & 9 & 4 \\
        Premature actuation & 6 & 1 & 9 & 2 & 7 & 2 \\
        \bottomrule
    \end{tabular}
    \end{adjustbox}
    \par\vspace{3pt}
    \begin{minipage}{\linewidth}
        \footnotesize\raggedright
        w/o / w/: without / with arm state. Premature actuation counts failed
        trials carrying that event label; successful trials are outside this
        annotation cohort.
    \end{minipage}
\endgroup
\end{figure}

Beaker Pouring shows holding and further actuation within one operation
(Fig.~\ref{fig:beaker_contact_phases}). Initial closure establishes a grasp;
limited width change maintains contact; further closure acts on the held
object before reopening ends the operation. Width and force traces show
how the gripper maintains contact and changes its interaction with the
object as the task progresses.

Bag Clipping shows contact observations evolving during object use
(Fig.~\ref{fig:clip_contact_observations}). Acquiring the clip, applying it
to the bag, and releasing it require successive contact operations.
Difference images from both fingertips change across these stages,
capturing changes at the gripper--object interface. Approximate phase
correspondence relates these observations to acquisition, application,
and release.

These executions show fine-grained control through target retention and
adjustment. The arm-state comparisons in Q3
(Fig.~\ref{fig:arm_state_ablation}, Table~\ref{tab:failure_patterns}) test
its coordination with task motion. ContactPolicy
implements this control through a persistent target and predicted
increments. Modulation selects whether to retain the target or apply an
adjustment while new contact observations are processed. Holding and
further actuation share a local feedback process, regulating contact as
the arm advances the task.

\subsection{Q3: Does Arm State Improve Coordination?}
\label{sec:arm_state_ablation}

Arm state coordinates local gripper regulation with the arm's task
progress. We remove arm state after temporal aggregation while retaining
tactile, force, and gripper-state inputs under the same training and
evaluation protocols. Adding arm state raises successful executions from
43/75 to 63/75 across the three tool-use tasks, with improvements on every
task (Fig.~\ref{fig:arm_state_ablation}, Table~\ref{tab:failure_patterns}).

\begin{figure}[!t]
\centering
    \includegraphics[width=\linewidth,trim=0bp 5bp 0bp 0.5bp,clip]{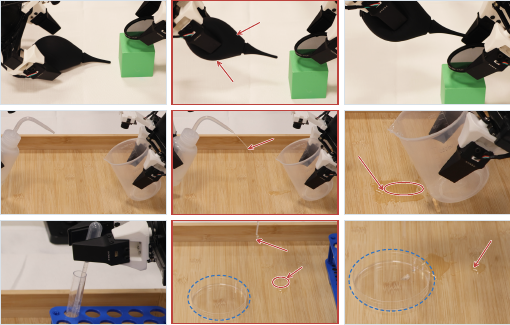}
    \caption{\textbf{Premature actuation during task motion.} Without arm-state input, the blower is squeezed flat before its nozzle reaches the lens; the beaker is squeezed before alignment with the receiving vessel is complete, causing premature water discharge; and liquid leaks from the dropper while it is being moved towards the dish.}
    \label{fig:premature_examples}
\par\vspace{6pt}
    \includegraphics[width=\linewidth]{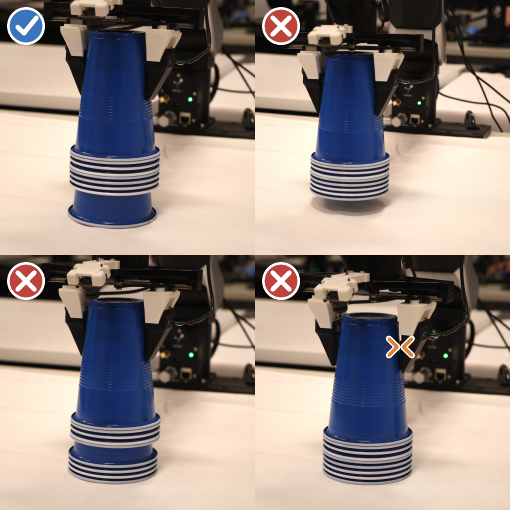}
    \caption{\textbf{Cup Separation outcomes.} Successful separation leaves one cup on the table. Failure cases show excessive gripping force that lifts the entire cup stack, insufficient force that leaves too many cups on the table, and failure to clamp the cup.}
    \label{fig:split_cups_cases}
\end{figure}

\label{sec:failure_analysis}
This improvement accompanies fewer failures involving premature actuation.
Total failures fall from 32 to 12, and failed trials carrying the premature
actuation label fall from 22 to 5 (Table~\ref{tab:failure_patterns}). Each
failed trial receives at most one such label.
Fig.~\ref{fig:premature_examples} shows the corresponding
behaviours: the blower is squeezed before its nozzle reaches the lens,
the beaker discharges water before alignment is complete, and the dropper
leaks while moving towards the dish. These examples show that grasping a tool must be followed by
actuation coordinated with its arrival at the intended target.

The three tasks share a coordination requirement: the gripper must adjust
to contact conditions while relating tool operation to the task stage
reached by the arm. Contact history describes how the grasp evolves, and
current arm state provides motion and position context for the next
adjustment. ContactPolicy uses this context for gripper-width control,
coordinating local contact regulation with ongoing arm execution. The gripper remains the controlled component, and arm state relates
each local adjustment to the ongoing progress of the manipulation task.

\subsection{Q4: Which Inputs Support Local Regulation?}
\label{sec:wrist_camera}

\begin{table}[!t]
\centering
    \small
    \tabletitle{\textbf{Wrist-camera ablation.} Task success rates (\%) and successful trials out of 25 per task, with the mean over four tasks.}
    \label{tab:wrist_camera}
    \begin{adjustbox}{max width=\linewidth}
    \begin{tabular}{@{}lcc@{}}
        \toprule
        \textbf{Task} & \textbf{Without wrist RGB} & \textbf{With wrist RGB} \\
        \midrule
        Tube Opening & 100\% (25/25) & 96\% (24/25) \\
        Bag Clipping & 92\% (23/25) & 84\% (21/25) \\
        Beaker Pouring & 80\% (20/25) & 64\% (16/25) \\
        Petri Dispensing & 84\% (21/25) & 80\% (20/25) \\
        \midrule
        Mean & 89\% & 81\% \\
        \bottomrule
    \end{tabular}
    \end{adjustbox}
\end{table}

The wrist-camera ablation tests whether adding visual observations inside
the local controller improves its contact decisions
(Table~\ref{tab:wrist_camera}). Wrist RGB is added to the temporal input
alongside tactile and force histories, while current arm/gripper state,
the visual base policy, and Intent switching are retained. Both configurations retain visual task guidance, isolating the addition
of wrist images to ContactPolicy.

The original input set completes 89/100 trials across the four tasks,
compared with 81/100 after adding wrist RGB, and performs better on each
task. The largest difference occurs in Beaker Pouring, where contact
adjustment must serve tool operation during arm motion. Combined with the
arm-state ablation, this result supports a focused division of sensing:
contact history describes the local interaction, current state relates it
to execution, and vision guides the surrounding task. The comparison supports this allocation of sensing roles within the
evaluated ContactPolicy system.

The ablations address complementary sensing roles: Q3 establishes the
benefit of arm-state context, while Q4 evaluates adding scene information
to the local input. Together, these comparisons support retaining
execution context in a controller focused on contact feedback.

\subsection{Q5: What Makes Contact Control Task-Aligned?}
\label{sec:remaining_failures}

Figs.~\ref{fig:premature_examples} and~\ref{fig:split_cups_cases} connect
the preceding results to two requirements: act at the appropriate stage
and produce the intended object configuration. In the tool-use examples
without arm state, the blower compresses before its nozzle reaches the
lens, pouring produces visible spillage, and dispensing starts before the
receiving dish is reached. Holding the tool already permits actuation,
but successful use requires that actuation to serve the task at that
moment. This explains why execution context matters even when the learned
local policy commands only the gripper.

Cup Separation makes the object-level requirement explicit. In the
check-marked outcome, exactly one cup remains on the table while the
gripper lifts the rest of the stack. Cross-marked outcomes include lifting
the entire stack and leaving multiple cups. The 13 successful trials with
ContactPolicy therefore represent controlled separation, extending the
role of gripper feedback beyond grasp acquisition. The unsuccessful
examples expose the distinction between establishing a grasp and
achieving the required one-cup separation.

This distinction also sharpens the meaning of the complete-task gains.
For tool use, a grasp must remain useful until the tool reaches the point
of operation. For nested cups, the same grasp must allow the intended
relative motion between objects. ContactPolicy addresses both requirements through feedback that adjusts
an established grasp to support the intended interaction between objects.

Rewind-IL detects failures and returns to verified safe intermediate states
before restarting policy inference~\cite{zheng2026rewindil}. Such recovery
could complement local contact regulation when adjustments to an
established grasp no longer suffice.

Across the five questions, the evidence supports the paper's central
claim: fine-grained gripper control is a task-local feedback problem.
Complete-task comparisons establish its practical benefit, execution
examples reveal the changing contact demands, and input ablations connect
local sensing to coordination. ContactPolicy addresses these demands by
learning how and when to update the gripper target while the visuomotor
policy continues to advance the task.

\section{Conclusion}

ContactPolicy learns fine-grained gripper control from contact histories and
robot state during visual execution. Real-robot results show improved task
success, while execution analysis and input ablations identify the roles of
contact stage, motion context, and local sensing. Visual task motion can
therefore remain with the base policy while a local policy handles contact-stage
gripper adjustment. Task progression and contact correction can use different
time scales. This interface may transfer to new contact-rich tasks.

\bibliographystyle{IEEEtran}
\bibliography{references,bibliography_style}
\end{document}